\documentclass[final]{MIT8X10M}
\slugtoks{Hazan, Papandreou, Tarlow: Perturbations, Optimization, and Statistics}

\usepackage{fleqn}    %% for compatibility with amsmath fleqn 
\usepackage{booktabs} %% professional looking tables
\usepackage{multirow} %% multiple row cells in tables
\usepackage{subfig}
\usepackage{subfloat}
\newcounter{subfigure@save}[figure]  % Hack to be able to use the subfloat package
\usepackage{color}
\usepackage{ifthen}
\usepackage{eucal}

\usepackage{graphicx}
\usepackage[sectionbib]{chapterbib_numbered_bibliography}
\usepackage[sectionbib,round]{natbib}

\usepackage[fleqn]{amsmath}
\usepackage{amssymb}
\usepackage{amsbsy}

\usepackage[chapter]{algorithm}
\usepackage{algorithmicx}
\usepackage{algpseudocode} %% loosli
\usepackage[ruled,vlined]{algorithm2e} %% haffner, raykar HACKED!

\let\proof\relax

\usepackage{amsthm}
\newtheorem{theorem}{Theorem}[chapter]

\newtheorem{assumption}{Assumption}[chapter]

\newtheorem{corollary}[theorem]{Corollary}

\newtheorem{remark}{Remark}[chapter]

\usepackage{url}

\usepackage{smallcaptions}

\usepackage{bm}

\usepackage{pdfpages}

\usepackage{titletoc}
\newcommand{\hc}{\mathcal{h}}

\newcommand{\qc}{\mathcal{q}}

\newcommand{\R}{\mathbb{R}}
\newcommand{\N}{\mathbb{N}}

\DeclareMathOperator*{\argmin}{argmin}

\newcommand{\Fig}[1]{Figure~\protect\ref{#1}}

\newcommand{\Sec}[1]{Section~\protect\ref{#1}}
\mathchardef\ordinarycolon\mathcode`\:
\mathcode`\:=\string"8000
\begingroup \catcode`\:=\active
  \gdef:{\mathrel{\mathop\ordinarycolon}}
\endgroup

\newenvironment{abstract}
 {\begin{list}{}
    {\setlength{\leftmargin}{0em}
     \setlength{\rightmargin}{0em}
     \setlength{\itemsep}{0pt}
     \setlength{\topsep}{0pt}} 
    \item[] \em }
 {\end{list}\par\vspace{8ex}\par}

\newenvironment{authors}
 {\vspace{-4ex}\begin{list}{}
    {\setlength{\leftmargin}{-.5\marginpartotal}
     \setlength{\rightmargin}{0pt}
     \setlength{\itemsep}{0pt}
     \setlength{\topsep}{0pt}} 
    \item[] }
 {\end{list}\par\vspace{8ex}\par}

\def\AUname#1{\par\makebox[2.8in][l]{\bf #1}}
\def\AUemail#1{{\tt #1}}
\def\AUweb#1{}%{\\{\sl #1}}
\def\AUaffiliation#1{\\\emph{#1}}
\def\AUaddress#1{\\\emph{#1}\par\medskip}

\makeatletter
\renewcommand{\small}{\@setfontsize\normalsize\@xpt\@xipt}
\renewcommand\appendix[1]{\par%
  \addtocounter{section}{1}%
  \setcounter{subsection}{0}%
  \section*{Appendix: #1}%
}
\makeatother

\numberwithin{equation}{chapter}
\begin{document}
% \bibliographystyle{abbrvnat}

%\fussy
\raggedbottom

% TITLE (UNOFFICIAL)
\pagestyle{empty}
%\cleardoublepage
%\chapter*{Advanced Structured Prediction}
%\begin{authors}
% 
%
%Editors:
%\par
%\bigskip
%
% \AUname{Tamir Hazan}
% \AUemail{tamir.hazan@technion.ac.il}
% \AUaffiliation{Technion - Israel Institute of Technology}
% \AUaddress{Technion City, Haifa 32000, Israel}
%
% \AUname{George Papandreou}
% \AUemail{gpapan@google.com}
% \AUaffiliation{Google Inc.}
% \AUaddress{340 Main St., Los Angeles, CA 90291 USA}
%
% \AUname{Daniel Tarlow}
% \AUemail{dtarlow@microsoft.com}
% \AUaffiliation{Microsoft Research}
% \AUaddress{Cambridge, CB1 2FB, United Kingdom}
%
%\end{authors}

%\vfill
%This is a draft version of the author chapter.
%Please check the formatting, editor changes, and your affiliation.
%Please use this version for sending us future modifications.
%\vfill

%\fontsize{11}{14}\selectfont
%\noindent
%The MIT Press\\
%Cambridge, Massachusetts\\
%London, England

% if preface ends on an odd-numbered page, insert a blank page between it and page 1
\cleardoublepage

\pagenumbering{arabic}
\pagestyle{normalheadings}
\setcounter{page}{1}

%%%%%%%%%%%%%%%%%%%%%%%%%%%%%%%%%%%%%%%%%%%%%%%%%%%%%%%%%%%%%%%%%%%%%
% CHAPTERS                                                          %
%%%%%%%%%%%%%%%%%%%%%%%%%%%%%%%%%%%%%%%%%%%%%%%%%%%%%%%%%%%%%%%%%%%%%

\begingroup

%%% Chapter-specific macros
% If you require additional macros beyond those in the main .tex file, please
% define them here only.
%%%
%%%
%%% (own macros after this line)

%\newcommand{\examplegf}{{\mathfrak{g}}}	% example chapter-specific macro
%\input lastname/std-macros

\theoremstyle{definition}
\newfloat{algbox}{tbp}{lop}
\newtheorem{alg}{Procedure}
\newcommand{\myalg}[3]{
\begin{center}
\fbox{
\parbox{0.95\textwidth}{
\begin{alg}\label{#1}{\textsc{ #2}}
\vspace{.1cm}\\ #3 %\emph{ #3}
\end{alg}
}}
\end{center}
}

\newcommand\BP{\ensuremath{\mathbb{P}}}
\newcommand\pl[1]{\textcolor{red}{[PL: #1]}}
\newcommand\wf[1]{\textcolor{blue}{[WF: #1]}}
\newcommand\sw[1]{\textcolor{green}{[WF: #1]}}

\newcommand{\Pois}{\text{Pois}}
\newcommand{\tT}{\widetilde{T}}

\newcommand\ShowFig[4]{\begin{figure}[ht] \begin{center} \includegraphics[scale=#2]{#1} \end{center} \caption{\label{fig:#3} #4} \end{figure}}
\newcommand\FigTop[4]{\begin{figure}[t] \begin{center} \includegraphics[scale=#2]{#1} \end{center} \caption{\label{fig:#3} #4} \end{figure}}
\newcommand\refsec[1]{Section~\ref{sec:#1}}
\newcommand\refeqn[1]{(\ref{eqn:#1})}
\newcommand\reffig[1]{Figure~\ref{fig:#1}}
\newcommand\eqdef{\ensuremath{\stackrel{\rm def}{=}}} % Equal by definition

\newcommand\hbetak{\hbeta^{(k)}}
\newcommand\hck{\hat c^{(k)}}
\newcommand\ck{c^{(k)}}
\newcommand\limloss{\rho}

\renewcommand\Xi{X^{(i)}}
\newcommand\Yi{Y^{(i)}}

\input lastname/stefan_macros

%%% (no further macros after this line)
%%%
%%%

\newcommand{\theTitle}{Data Augmentation via L\'evy Processes}
%\newcommand{\theTitle}{Rewinding L\'evy Processes for Data Augmentation}

%%% Running chapter title at the top of the page
\chapter[\theTitle \\
{\normalsize\rm\emph{A.~Example, B.~Author, and C.~Author}}]%
{\theTitle}

%%% The following label must be defined as chapter:firstauthorlastname
\label{chapter:example}

% TODO: explain long/short chapter titles
\markboth{\theTitle}{}

%%% Author list: please provide full details as in this example
\begin{authors}
\AUname{Stefan Wager}
\AUemail{swager@stanford.edu}
\AUaffiliation{Stanford University}
\AUaddress{Stanford, USA}
\AUname{William Fithian}
\AUemail{wfithian@berkeley.edu}
\AUaffiliation{University of California, Berkeley}
\AUaddress{Berkeley, USA} 
\AUname{Percy Liang}
\AUemail{pliang@cs.stanford.edu}
\AUaffiliation{Stanford University}
\AUaddress{Stanford, USA}
\end{authors}

\begin{abstract}
If a document is about travel, we may expect that
short snippets of the document should also be about travel.
We introduce a general framework for incorporating these types of invariances
into a discriminative classifier.
The framework imagines data as being drawn from a slice of a L\'evy process.
If we slice the L\'evy process at an earlier point in time, we obtain additional
pseudo-examples, which can be used to train the classifier.
We show that this scheme has two desirable properties:
it preserves the Bayes decision boundary,
and it is equivalent to fitting a generative model in the limit where we rewind time back to 0.
Our construction captures popular schemes such as
Gaussian feature noising and dropout training, as well as admitting new generalizations.
\end{abstract}

\section{Introduction}

% Feature noising of discriminative classifiers
Black-box discriminative classifiers such as logistic regression, neural networks,
and SVMs are the go-to solution in machine learning: they are simple to
apply and often perform well. However, an expert may have additional
knowledge to exploit, often taking the form of a certain family of transformations that should
usually leave labels fixed. For example, in object recognition,
an image of a cat rotated, translated, and peppered with a small amount of noise is probably still a cat.
Likewise, in document classification,
the first paragraph of an article about travel is most likely still about travel. In both cases, the
``expert knowledge'' amounts to a belief that a certain transform of the features should generally not affect
an example's label.

One popular strategy for encoding such a belief is {\em data augmentation}: generating additional
pseudo-examples or ``hints'' by applying label-invariant transformations to training examples' features
\citep{abu1990learning,scholkopf1997improving,simard1998transformation}. That is, each example $(X^{(i)}, Y^{(i)})$ is
replaced by many pairs $(\tX^{(i,b)}, Y^{(i)})$ for $b=1,\ldots, B$, where each $\tX^{(i,b)}$ is a transformed version of $X^{(i)}$.
This strategy is simple and modular: after generating the pseudo-examples, we can simply apply any supervised 
learning algorithm to the augmented dataset. Figure~\ref{fig:noisingExamples} illustrates two examples of this approach, 
an image transformed to a noisy image and a text caption, transformed by deleting words.

\FigTop{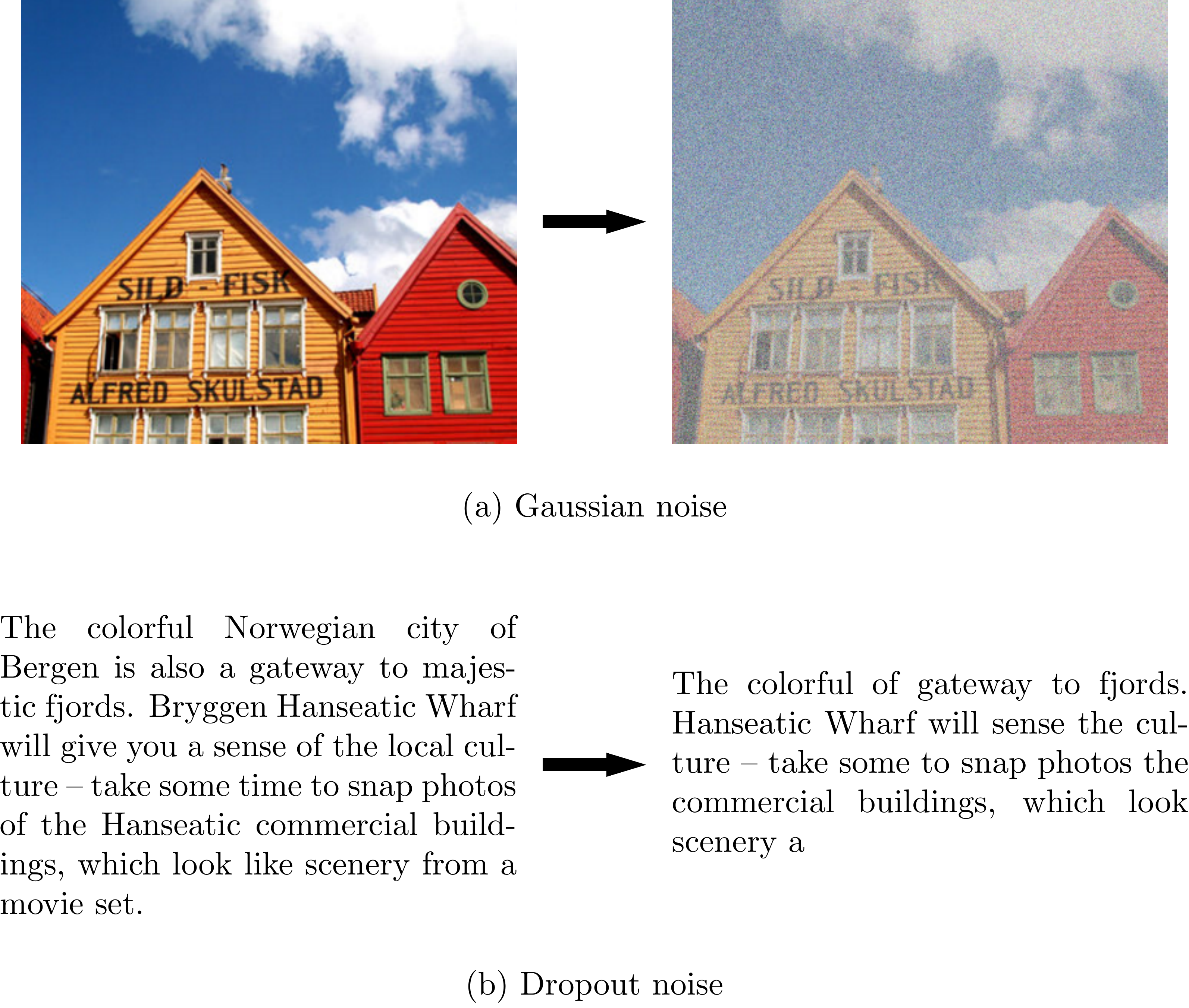}{0.35}{noisingExamples}{
Two examples of transforming an original input $X$ into a noisy, less
informative input $\tX$.  The new inputs clearly have the same label but
contain less information and thus are harder to classify.
}

Dropout training \citep{srivastava2014dropout} is an instance of data augmentation
that, when applied to an input feature vector, zeros out a subset of the features randomly.
Intuitively, dropout implies a certain amount of signal redundancy across features---that
an input with about half the features masked should usually be classified the same way as a fully-observed input.
In the setting of document classification, dropout can be seen as creating pseudo-examples
by randomly omitting some information (i.e., words) from each training example. Building
on this interpretation, \citet{wager2014altitude} show that learning with such artificially
difficult examples can substantially improve the generalization performance of a classifier.

% Characetrizing dropout with documents
To study dropout, \citet{wager2014altitude} assume that documents can be summarized
as Poisson word counts. Specifically, assume that each document has an underlying topic associated
with a word frequency distribution $\pi$ on the $p$-dimensional simplex and an expected length $T \geq 0$, 
and that, given $\pi$ and $T$, the word counts $X_j$ are independently
generated as $X_j \cond T, \, \pi \sim \Pois(T\,\pi_j)$.
The analysis of \citet{wager2014altitude} then builds on a duality between dropout and the
above generative model. Consider the example given in Figure \ref{fig:noisingExamples}, where dropout creates
pseudo-documents $\tX$ by deleting half the words at random from the original document $X$.
As explained in detail in Section \ref{sec:motivation}, if $X$ itself is drawn from the above
Poisson model, then the dropout pseudo-examples $\tX$ are marginally distributed 
as $\tX_j \cond T, \, \pi \sim \Pois(0.5 \,T\,\pi_j)$.
Thus, in the context of this Poisson generative model, dropout enables us to create new, shorter
pseudo-examples that preserve the generative structure of the problem.

% General Levy processes
The above interpretation of dropout raises the following question: if feature deletion is
a natural way to create information-poor pseudo-examples for document classification, are there
natural analogous feature noising schemes that can be applied to other problems?
In this chapter, we seek to address this question, and study a more general family of data augmentation
methods generalizing dropout, based on L\'evy processes:
We propose an abstract L\'evy thinning scheme that reduces to dropout in the
Poisson generative model considered by \citet{wager2014altitude}.
Our framework further suggests new methods for feature noising
such as Gamma noising based on alternative generative models,
all while allowing for a unified theoretical analysis.
\

\paragraph{From generative modeling to data augmentation.}

In the above discussion, we treated the expected document length $T$ as fixed. More generally, 
we could imagine the document as growing in length over time, with the observed document $X$ 
merely a ``snapshot'' of what the document looks like at time $T$. Formally, we can imagine a latent
Poisson process $(A_t)_{t \ge 0}$, with fixed-$t$ marginals $(A_t)_j \cond \pi \sim \Pois(t \, \pi_j)$,
and set $X = A_T$. In this notation, dropout amounts to ``rewinding'' the process 
$A_t$ to obtain short pseudo-examples.  By setting $\tX = A_{\alpha T}$, we have
$\BP[\tX = \tx \cond X=x] = \BP[A_{\alpha T} = \tx \cond A_T=x]$, 
for thinning parameter $\alpha\in (0,1)$.

The main result of this chapter is that the analytic tools developed by
\citet{wager2014altitude} are not restricted to the case where $(A_t)$ is a
Poisson process, and in fact hold whenever $(A_t)$ is a \emph{L\'evy process}.
In other words, their analysis applies to any classification problem where the
features $X$ can be understood as time-$T$ snapshots of a process $(A_t)$, i.e., $X = A_T$.

Recall that a L\'evy process $(A_t)_{t \ge 0}$ is a stochastic process with $A_0 = 0$
that has independent and stationary increments:
$\{ A_{t_i} - A_{t_{i-1}} \}$ are independent for $0 = t_0 < t_1 < t_2 < \cdots$,
and $A_t - A_s \eqd A_{t-s}$ for and $s < t$.
Common examples of L\'evy processes include Brownian motion and Poisson processes.

\FigTop{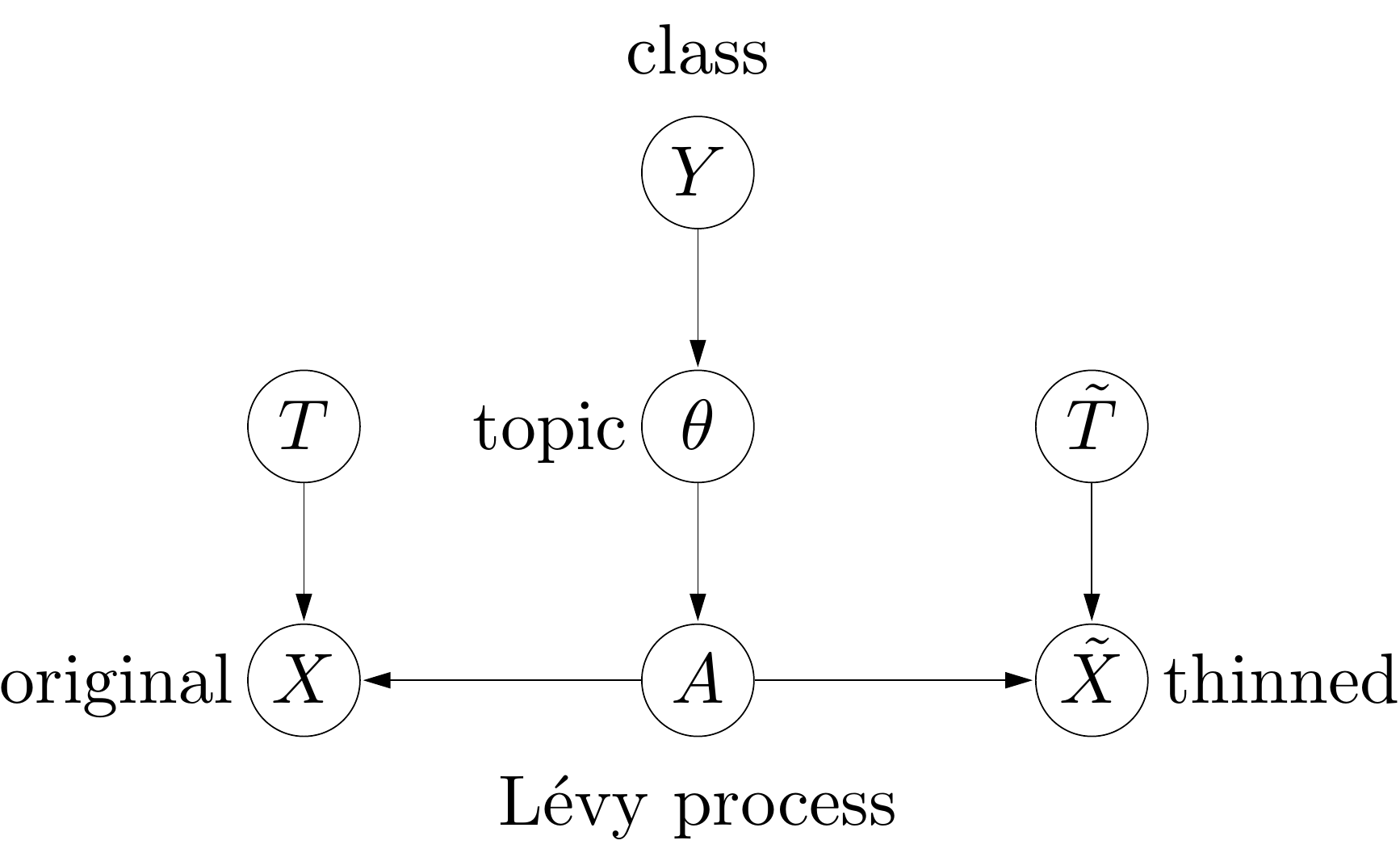}{0.4}{graphicalModel}{Graphical model
  depicting our generative assumptions; note that we are not fitting this generative model.
  Given class $Y$, we draw a topic $\theta$, which governs the parameters of the L\'evy process $(A_t)$.
  We slice at time $T$ to get the original input $X = A_T$ and at an earlier time
  $\tilde T$ to get the thinned or noised input $\tilde X = A_{\tilde T}$.
  We show that given $X$, we can sample $\tilde X$ without knowledge of $\theta$.
}

In any such L\'evy setup, we show that it is possible to devise an analogue to dropout
that creates pseudo-examples by \emph{rewinding} the process back to some earlier time
$\tT \leq T$.
Our generative model is depicted in \reffig{graphicalModel}: $(A_t)$, the information
relevant to classifying $Y$, is governed by a latent topic $\theta \in \RR^p$.
L\'evy thinning then seeks to rewind $(A_t)$---importantly as we shall see, without having access to $\theta$.

We should think of $(A_t)$ as representing an ever-accumulating amount of information
concerning the topic $\theta$: In the case of document classification, $(A_t)$ are
the word counts associated with a document that grows longer as $t$ increases.
In other examples that we discuss in Section~\ref{sec:examples}, 
$A_t$ will represent the sum of $t$ independent noisy sensor readings. The independence of increments 
property assures that as we progress in time, we are always obtaining new information.
The stopping time $T$ thus represents the \emph{information content} in input $X$ about topic $\theta$.
L\'evy thinning seeks to improve classification accuracy by turning a few
information-rich examples $X$ into many information-poor examples $\tX$.

\FigTop{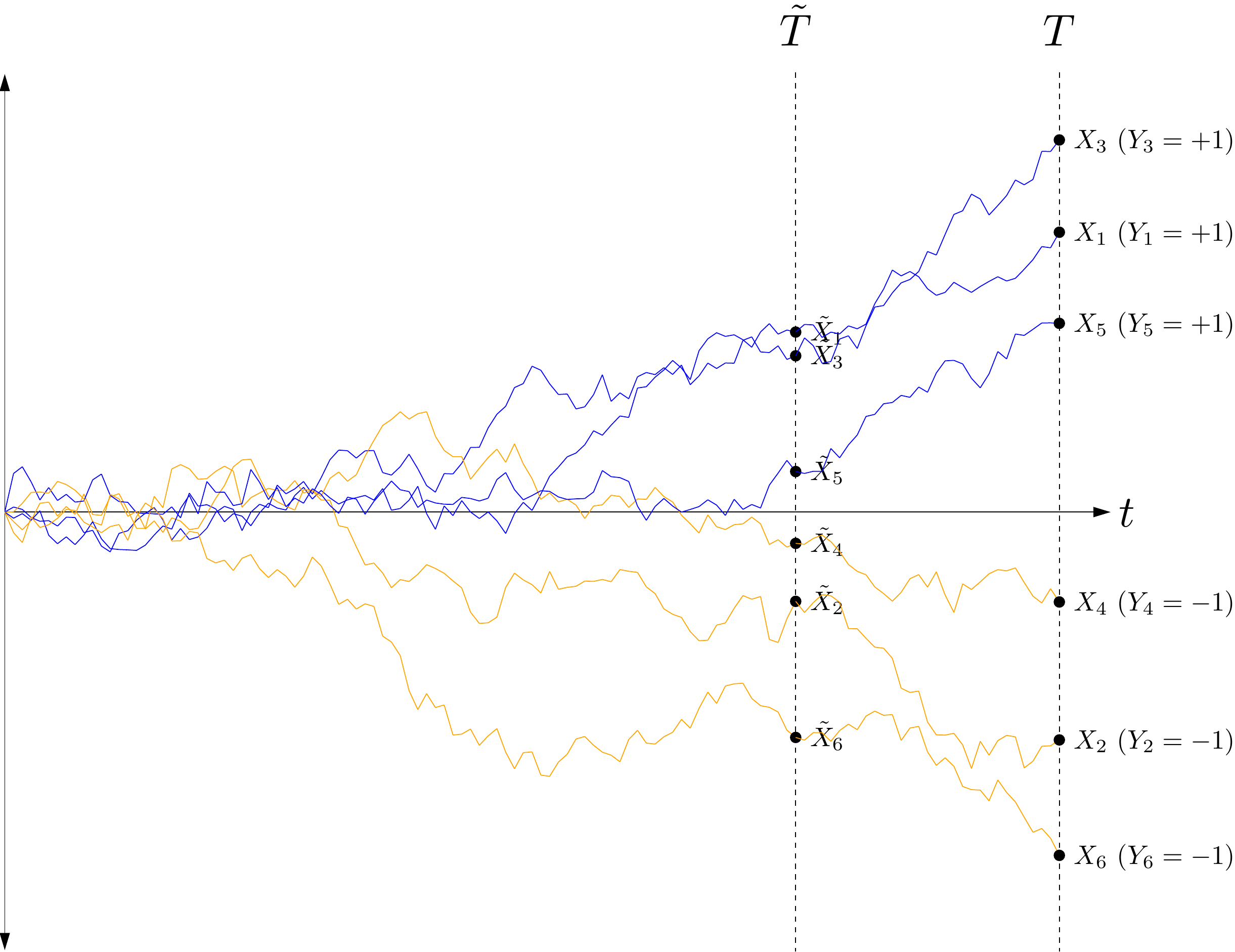}{0.4}{levyProcess}{
We model each input $X$ as a slice of a L\'evy process at time $T$.
We generate noised examples $\tX$ by ``stepping back in time'' to $\tilde T$.
Note that the examples of the two classes are closer together now,
thus forcing the classifier to work harder.
}

We emphasize that, although our approach uses generative modeling
to motivate a data augmentation scheme, we do not in fact fit a
generative model. This presents a contrast to the prevailing practice:
two classical approaches to multiclass classification are to either 
directly train a discriminative model by running, e.g., multiclass logistic regression on
the $n$ original training examples; or, at the other extreme, to
specify and fit a simple parametric version of the above generative model, e.g., naive Bayes,
and then use Bayes' rule for classification.
It is well known that the latter approach is usually more efficient if it has access to a correctly
specified generative model, but may be badly biased in case of model misspecification
\citep{efron1975efficiency,ng02compare,liang08asymptotics}.
Here, we first seek to devise a noising scheme $X \rightarrow \tX$ and then to train a discriminative
model on the pseudo-examples $(\tX, \, Y)$ instead of the original examples $(X, \, Y)$. Note that even
if the generative model is incorrect, this approach will incur limited bias as long as the noising scheme
roughly preserves class boundaries --- for example, even if the Poisson document model is incorrect, 
we may still be justified in classifying a subsampled travel document as a travel document. 
As a result, this approach can take advantage of an abstract generative structure
while remaining more robust to model misspecification than parametric generative modeling.
\

\paragraph{Overview of results.}

We consider the multiclass classification setting
where we seek to estimate a mapping from input $X$ to class label $Y$.
We imagine that each $X$ is generated by a mixture of L\'evy process,
where we first draw a random topic $\theta$ given the class $Y$, and then
run a L\'evy process $(A_t)$ depending on $\theta$ to time $T$.
In order to train a classifier, we pick a thinning parameter $\alpha \in (0, \, 1)$,
and then create pseudo examples by rewinding the original $X$
back to time $\alpha \, T$, i.e., $\tX \sim A_{\alpha T} \cond A_T$.

We show three main results in this chapter.
Our first result is that we can generate such pseudo-examples $\tX$
without knowledge of the parameters $\theta$ governing the generative L\'evy process.
In other words, while our method posits the existence of a generative model,
our algorithm does not actually need to estimate it.
Instead, it enables us to give hints about a potentially complex generative structure
to a discriminative model such as logistic regression.

Second, under assumptions that our generative model is correct,
we show that feature noising preserves the Bayes decision boundary:
$\BP[Y \mid X = x] = \BP[Y \mid \tX = x]$.
This means that feature noising does not introduce any bias in the infinite data limit.

Third, we consider the limit of rewinding to the beginning of time ($\alpha \to 0$).
Here, we establish conditions given which, even with finite data,
the decision boundary obtained by fitting a linear classifier on the pseudo-examples is equivalent
to that induced by a simplified generative model. When this latter result holds,
we can interpret $\alpha$-thinning as providing a semi-generative regularization
path for logistic regression, with a simple generative procedure at one end and
unregularized logistic regression at the other.
\

\paragraph{Related work.}

% Generative versus discriminative
The trade-off between generative models and discriminative models has been
explored extensively.  \citet{rubinstein1997discriminative} empirically compare
discriminative and generative classifiers models with respect to bias and variance,
\citet{efron1975efficiency} and \citet{ng02compare} provide a more formal
discussion of the bias-variance trade-off between logistic regression and naive Bayes.
\citet{liang08asymptotics} perform an asymptotic analysis for general exponential families.

A number of papers study hybrid loss functions that combine both a joint and conditional likelihood
\citep{raina04hybrid,bouchard04tradeoff,lasserre06hybrid,pal06mcl,liang08asymptotics}.
The data augmentation approach we advocate in this chapter is fundamentally different,
in that we are merely using the structural assumptions implied by the generative models
to generate more data, and are not explicitly fitting a full generative model.

% Dropout
The present work was initially motivated by understanding
dropout training \citep{srivastava2014dropout},
which was introduced in the context of regularizing deep neural networks,
and has had much empirical success
\citep{ba2013adaptive,goodfellow2013maxout,krizhevsky2012imagenet,wan2013regularization}.
Many of the regularization benefits of dropout
can be found in logistic regression and other single-layer models,
where it is also known as ``blankout noise'' \citep{globerson2006nightmare,maaten2013learning}
and has been successful in natural language tasks such as document classification and named entity recognition
\citep{wager2013dropout,wang2013fast,wang2013noising}.
There are a number of theoretical analyses of dropout:
using PAC-Bayes framework \citep{mcallester2013pac},
comparing dropout to ``altitude training'' \citep{wager2014altitude},
and interpreting dropout as a form of adaptive regularization \citep{baldi2014dropout,bishop1995training,helmbold2015inductive,josse2014stable,wager2013dropout}.

\section{L\'evy Thinning}

We begin by briefly reviewing the results of \citet{wager2014altitude}, who study
dropout training for document classification from the perspective of thinning documents (\refsec{motivation}).
Then, in \refsec{setup}, we generalize these results to the setting
of generic L\'evy generative models.

\subsection{Motivating Example: Thinning Poisson Documents}\label{sec:motivation}

Suppose we want to classify documents according to their subject, e.g., sports, politics, or travel.
As discussed in the introduction, common sense intuition about the nature of documents
suggests that a short snippet of a sports document should also be classified as a sports document.
If so, we can generate many new training examples by cutting up the original documents in
our dataset into shorter subdocuments and labeling each subdocument with the same label
as the original document it came from. By training a classifier on all of the pseudo-examples
we generate in this way, we should be able to obtain a better classifier.

In order to formalize this intuition, we can represent a document as a sequence of words from
a dictionary $\{1, \, \ldots, \, d\}$, with the word count $X_{j}$ denoting the number of
occurrences of word $j$ in the document. Given this representation, we can easily create
``subdocuments'' by binomially downsampling the word counts $X_j$ independently. That is,
for some fixed downsampling fraction $\alpha\in (0,1)$, we draw
\begin{align}
\tX_j \mid X_j \sim \text{Binom}(X_j, \alpha).
\end{align}
In other words, we keep each occurrence of word $j$ independently with probability $\alpha$.

\citet{wager2014altitude} study this downsampling scheme in the context of a
Poisson mixture model for the inputs $X$ that obeys the structure of \reffig{graphicalModel}:
first, we draw a class $Y \in \{1, \dots, K\}$ (e.g., travel) and a ``topic'' $\theta \in \R^d$
(e.g., corresponding to travel in Norway).
The topic $\theta$ specifies a distribution over words,
\begin{align}
  \mu_j(\theta) = e^{\theta_j},
\end{align}
where, without loss of generality, we assume that $\sum_{j=1}^d e^{\theta_{j}} = 1$.
We then draw a $\Pois(T)$ number of words, where $T$ is the expected document length,
and generate each word independently according to $\theta$.
Equivalently, each word count is an independent Poisson random variable,
 $X_j \sim \Pois(T \mu_j(\theta))$.
The following is an example draw of a document:
\begin{align*}
  Y & = \text{travel} \\
  \theta & = [\overbrace{0.5}^\text{norway}, \overbrace{0.5}^\text{fjord}, \overbrace{1.2}^\text{the}, \overbrace{-2.7}^\text{skyscraper}, \dots] \\
  X & = [\overbrace{2}^\text{norway}, \overbrace{1}^\text{fjord}, \overbrace{3}^\text{the}, \overbrace{0}^\text{skyscraper}, \dots] \\
  \tX & = [\overbrace{1}^\text{norway}, \overbrace{0}^\text{fjord}, \overbrace{1}^\text{the}, \overbrace{0}^\text{skyscraper}, \dots] \\
\end{align*}
Let us now try to understand the downsampling scheme $\tX \mid X$ in the
context of the Poisson topic model over $X$.
For each word $j$, recall that $\tX_j \mid X_j \sim \text{Binom}(X_j, \alpha)$.
If we marginalize over $X$, then we have:
\begin{align}
  \tX_j \mid T, \, \theta \sim \text{Pois}\p{\alpha T \mu_j(\theta)}.
\end{align}
As a result, the distribution of $\tX$ is
exactly the distribution of $X$ if we replaced $T$ with $\tT = \alpha T$.

We can understand this thinning by embedding the document $X$ in a 
multivariate Poisson process $(A_t)_{t\ge 0}$, where the marginal distribution of 
$A_t \in \{0, 1, 2, \dots\}^d$ is defined to be the distribution over counts 
when the expected document length is $t$. Then, we can write
\begin{align}
  X = A_T, \quad \tX = A_{\tT}.
\end{align}
Thus, under the Poisson topic model,
the binomial thinning procedure does not
alter the structure of the problem other than by shifting the expected document length from $T$ to $\tT$.
\Fig{fig:pois-proc} illustrates one realization of L\'evy thinning in the Poisson
case with a three-word dictionary. Note that in this case we can sample $\tX = A_{\alpha T}$ given $X = A_T$ 
without knowledge of $\theta$.

This perspective lies at the heart of the analysis in \citet{wager2014altitude},
who show under the Poisson model that, when the overall document length $\|X\|_1$
is independent of the topic $\theta$,
thinning does not perturb the optimal decision boundary.
Indeed, the conditional distribution over class labels is identical for the original features
and the thinned features:
\begin{align}
  \label{eqn:labelPreserved}
  \BP[Y \mid X=x] = \BP[Y \mid \tX=x].
\end{align}
This chapter extends the result to general L\'evy processes (see Theorem
\ref{theo:bayes_boundary}).

This last result \refeqn{labelPreserved} may appear quite counterintuitive:
for example, if $A_{60}$ is more informative than $A_{40}$,
how can it be that downsampling does not perturb the
conditional class probabilities?
Suppose $x$ is a 40-word document ($\|x\|_1 = 40$).
When $t=60$, most of the documents will be longer than 40 words,
and thus $x$ will be less likely under $t = 60$ than under $t = 40$.
However, \refeqn{labelPreserved} is about the distribution of $Y$
\emph{conditioned on a particular realization} $x$.
The claim is that, having observed $x$, we obtain the same information about $Y$ regardless of
whether $t$, the expected document length, is $40$ or $60$.

%%%%%%%%%%%%%%%%%%%%%%%%%%%%%%%%%%%%%%%%%%%%%%%%%%%%%%%%%%%%
\subsection{Thinning L\'evy Processes}
\label{sec:setup}

The goal of this section is to extend the 
Poisson topic model from \Sec{sec:motivation} and construct general
thinning schemes with the invariance property of \refeqn{labelPreserved}.
We will see that L\'evy processes provide a
natural vehicle for such a generalization: The Poisson process used to generate
documents is a specific L\'evy process, and binomial sampling corresponds to
``rewinding'' the L\'evy process back in time.

Consider the multiclass classification problem of predicting a discrete
class $Y \in \{1, \dots, K\}$ given an input vector $X \in \R^d$.
Let us assume that the joint distribution over $(X, Y)$ is governed by the
following generative model:
\begin{enumerate}
\item Choose $Y \sim \text{Mult}(\pi)$, where $\pi$ is on the $K$-dimensional simplex.
\item Draw a topic $\theta \mid Y$, representing a subpopulation of class $Y$.
\item Construct a L\'evy process $(A_t)_{t \ge 0} \mid \theta$, where $A_t \in
  \R^d$ is a potential input vector at time $t$.
\item Observe the input vector $X = A_T$ at a fixed time $T$.
\end{enumerate}

% Generative model
While the L\'evy process imposes a fair amount of structure, we make no
assumptions about the number of topics, which could be uncountably infinite, or
about their distribution, which could be arbitrary.  Of course, in such an unconstrained
non-parametric setting, it would be extremely difficult to adequately fit the generative model.
Therefore, we take a different tack: We will use
the structure endowed by the L\'evy process to generate pseudo-examples
for consumption by a discriminative classifier.
These pseudo-examples implicitly encode our generative assumptions.

The natural way to generate a pseudo-example $(\tX, Y)$ is to ``rewind''
the L\'evy process $(A_t)$ backwards from time $T$ (recall $X = A_T$)
to an earlier time $\tT = \alpha T$ for some $\alpha \in (0, 1)$
and define the thinned input as $\tX = A_{\tT}$.
In practice, $(A_t)$ is unobserved, so we draw $\tX$ conditioned
on the original input $X = A_T$ and topic $\theta$.
In fact, we can draw many realizations of $\tX \mid X, \theta$.

Our hope is that a single full example $(X,Y)$ is rich enough
to generate many different pseudo-examples $(\tX, Y)$,
thus increasing the effective sample size.
Moreover, \citet{wager2014altitude} show
that training with such pseudo-examples can also lead to a somewhat surprising ``altitude training''
phenomenon whereby thinning yields an improvement in generalization
performance because the pseudo-examples are more difficult to classify than the
original examples, and thus force the learning algorithm to work harder and
learn a more robust model.

A technical difficulty is that generating $\tX \mid X, \theta$ seemingly requires
knowledge of the topic $\theta$ driving the underlying L\'evy process $(A_t)$.
In order to get around this issue,
we establish the following condition under which the observed input $X = A_T$ alone is sufficient---that is,
$\BP[\tX \mid X, \theta]$ does not actually depend on $\theta$.

\begin{assumption}[exponential family structure]
\label{assu:expfam}
The L\'{e}vy process $(A_t) \cond \theta$ is drawn according to an exponential family model 
whose marginal density at time $t$ is
\begin{equation}
\label{eq:assu1}
 f_{\theta}^{(t)}(x) = \exp\sqb{\theta \cdot x - t\psi\p{\theta}} h^{(t)}\p{x} \text{ for every } t \in \RR.
\end{equation}
Here, the topic $\theta \in \RR^d$ is an unknown parameter vector,
and $h^{(t)}(x)$ is a family of carrier densities indexed by $t \in \RR$.
\end{assumption}

The above assumption is a natural extension of a standard exponential family assumption
that holds for a single value of $t$. Specifically, suppose that $h^{(t)}(x)$, $t > 0$, denotes the $t$-marginal densities
of a L\'evy process, and that $f^{(1)}_\theta(x) = \exp\sqb{\theta \cdot x - \psi\p{\theta}} h^{(1)}(x)$ is an
exponential family through $h^{(1)}(x)$ indexed by $\theta \in \RR^d$. Then, we can verify that the densities
specified in \eqref{eq:assu1} induce a family of L\'evy processes indexed by $\theta$. The
key observation in establishing this result is that, because $h^{(t)}(x)$ is the $t$-marginal of a L\'evy
process, the L\'evy--Khintchine formula implies that
$$ \int e^{\theta \cdot x} h^{(t)}(x) \ dx = \p{ \int e^{\theta \cdot x}  h^{(1)}(x) \ dx }^t = e^{t \, \psi\p{\theta}}, $$
and so the densities in \eqref{eq:assu1} are properly normalized.

We also note that, given this assumption and as $T\to\infty$,
we have that ${A_T}/{T}$ converges almost surely to $\mu(\theta) \eqdef \EE{A_1}$.
Thus, the topic $\theta$ can be understood as a description of an infinitely informative
input. For finite values of $T$, $X$ represents a noisy observation of the topic $\theta$.

\begin{figure}
\centering
\includegraphics[width=\textwidth]{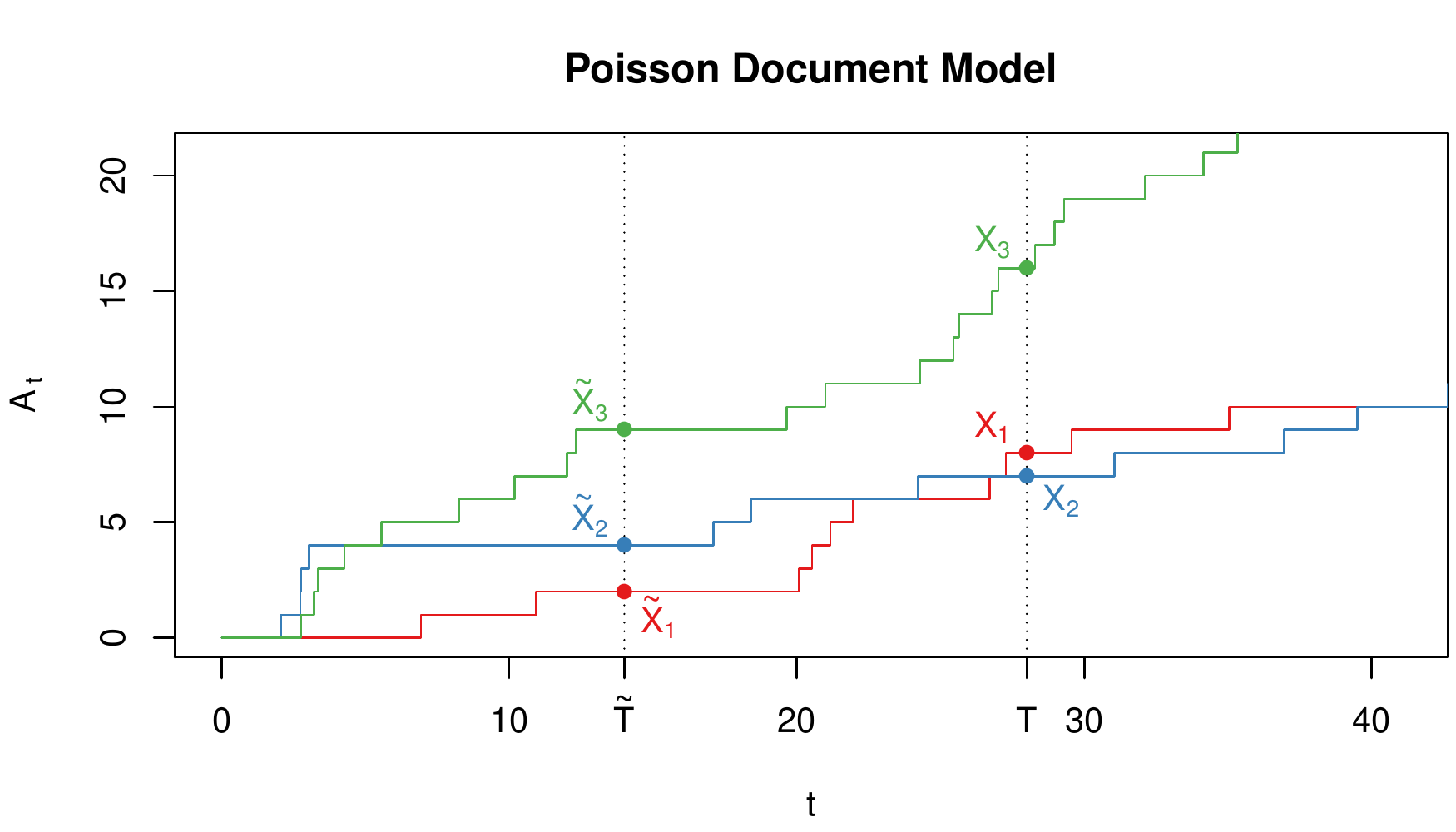}
\caption{Illustration of our Poisson process document model with a three-word dictionary and $\mu(\theta)=(0.25, 0.3, 0.45)$. The word counts of the original document, $X=(8,7,16)$, represents the trivariate Poisson process $A_t$, sliced at $T=28$. The thinned pseudo-document $\tX=(2,4,9)$ represents $A_t$ sliced at $\tT=14$.}\label{fig:pois-proc}
\end{figure}

Now, given this structure, we show that the distribution of $\tX = A_{\alpha T}$
conditional on $X = A_T$ does not depend on $\theta$.
Thus, feature thinning is possible without
knowledge of $\theta$ using the L\'evy thinning procedure defined below.
We note that, in our setting, the carrier distributions $h^{(t)}(x)$ are always known;
in Section \ref{sec:examples}, we discuss how to efficiently sample from the
induced distribution $g^{(\alpha T)}$ for some specific cases of interest.

\begin{theorem}[L\'evy thinning]
\label{defi:thinning}
\label{theo:marginal_thinning}
Assume that $(A_t)$ satisfies the exponential family structure in \eqref{eq:assu1}, and let
$\alpha \in (0, \, 1)$ be the thinning parameter.
Then, given an input $X = A_T$ and conditioned on any $\theta$,
the thinned input $\tX = A_{\alpha T}$ has the following density:
\begin{align}
\label{eq:thinning}
g^{(\alpha T)}(\tx; X) = \frac{h^{(\alpha T)}(\tx) \, h^{((1 - \alpha) T)}(X - \tx)}{h^{(T)}(X)},
\end{align}
which importantly does not depend on $\theta$.
\end{theorem}
\begin{proof}
  Because the L\'evy process $(A_t)$ has independent and stationary increments,
  we have that $A_{\alpha T} \sim f_\theta^{(\alpha T)}$
  and $A_{T} - A_{\alpha T} \sim f_\theta^{((1-\alpha) T)}$ are independent.
  Therefore, we can write the conditional density of $A_{\alpha T}$ given $A_T$
  as the joint density over $(A_{\alpha T}, A_T)$
  (equivalently, the reparametrization $(A_{\alpha T}, A_T - A_{\alpha T})$)
  divided by the marginal density over $A_T$:
\begin{align}
  g^{(\alpha T)}(\tx; X)
  &= \frac{f_\theta^{(\alpha T)}(\tx) f_\theta^{((1-\alpha)T)}(X - \tx)}{f_\theta^{(T)}(X)} \\
  &= \left(\exp\sqb{\theta \cdot \tx - \alpha T \psi(\theta)} h^{(\alpha T)}(\tx) \right) \nonumber \\
  & \times \left(\exp\sqb{\theta \cdot (X - \tx) - (1-\alpha) T \psi(\theta)} h^{((1-\alpha) T)}(X - \tx) \right) \nonumber \\
  & \times \left(\exp\sqb{\theta \cdot X - T \psi(\theta)} h^{(T)}(X) \right)^{-1}, \nonumber
\end{align}
where the last step expands everything \eqref{eq:assu1}.
Algebraic cancellation, which removes all dependence on $\theta$, completes the proof.
\end{proof}

Note that while Theorem~\ref{theo:marginal_thinning} guarantees we can carry out feature thinning
without knowing the topic $\theta$, it does not guarantee that we can do it without knowing the information
content $T$. For Poisson processes, the binomial thinning mechanism depends only on $\alpha$ and not on
the original $T$. This is a convenient property in the Poisson case but does not carry over to all L\'{e}vy
processes --- for example, if $B_t$ is a standard Brownian motion, then the distribution of $B_2$ given
$B_4=0$ is $\mathcal N(0,1)$, while the distribution of $B_{200}$ given $B_{400}=0$ is $\mathcal N(0,100)$. As we will see in Section~\ref{sec:examples}, thinning in the Gaussian and Gamma families does require knowing $T$,
which will correspond to a ``sample size'' or ``precision.'' Likewise, Theorem~\ref{theo:marginal_thinning} does
not guarantee that sampling from~\eqref{eq:thinning} can be carried out efficiently; however, in all the examples we present here,
sampling can be carried out easily in closed form.

%%%%%%%%%%%%%%%%%%%%%%%%%%%%%%%%%%%%%%%%%%%%%%%%%%%%%%%%%%%%
\subsection{Learning with Thinned Features}
\label{sec:learning}

\begin{algbox}[t]
\myalg{proc:levy_thinning}{Logistic Regression with L\'evy Regularization}{
Input: $n$ training examples $(X^{(i)}, Y^{(i)})$,
a thinning parameter $\alpha \in (0, \, 1)$,
and a feature map $\phi : \R^d \mapsto \R^p$.

\begin{enumerate}
  \item For each training example $X^{(i)}$,
    generate $B$ thinned versions $(\tX^{(i,b)})_{b=1}^B$
    according to \eqref{eq:thinning}.
\item Train logistic regression on the resulting pseudo-examples:
\begin{equation}
\label{eq:balpha}
\hbeta \eqdef \argmin_{\beta \in \R^{p \times K}} \cb{\sum_{i = 1}^n \sum_{b = 1}^B \ell\p{\beta; \, \tX^{(i,b)}, \, Y^{(i)}}},
\end{equation}
where the multi-class logistic loss with feature map $\phi$ is
\begin{align}
  \ell(\beta; x, y) \eqdef
  \log\p{\sum_{k=1}^K e^{\beta^{(k)} \cdot \phi(x)}}
  - \beta^{(y)} \cdot \phi(x).
\end{align}
\item Classify new examples according to
\begin{equation}
\label{eq:classif_calib}
\hy(x) = \argmin_{k \in \cb{1, \, \dots, \, K}} \cb{
  \hck - \hbetak \cdot \phi(x)},
\end{equation}
where the $\hatc_k \in \RR$ are optional class-specific calibration parameters for $k = 1, \dots, K$.
\end{enumerate}
}
\end{algbox}

Having shown how to thin the input $X$ to $\tX$ without knowledge of
$\theta$, we can proceed to defining our full data augmentation strategy.
We are given $n$ training examples $\{(X^{(i)}, Y^{(i)})\}_{i=1}^n$.
For each original input $X^{(i)}$, we generate $B$ thinned versions
$\tX^{(i,1)}, \dots, \tX^{(i,B)}$ by sampling from \eqref{eq:thinning}.
We then pair these $B$ examples up with $Y^{(i)}$ and train 
\emph{any} discriminative classifier on these $Bn$ examples.
Algorithm~\ref{proc:levy_thinning} describes the full procedure
where we specialize to logistic regression.
If one is implementing this procedure using stochastic gradient descent,
one can also generate a fresh thinned input $\tX$ whenever we sample an input $X$
on the fly, which is the usual implementation of dropout training \citep{srivastava2014dropout}.

In the final step \eqref{eq:classif_calib} of Algorithm~\ref{proc:levy_thinning}, we
also allow for class-specific calibration parameters $\hc$. After the $\hbetak$ have been
determined by logistic regression with L\'evy regularization, these parameters $\hck$
can be chosen by optimizing the logistic loss on the original uncorrupted training data.
As discussed in Section \ref{sec:end}, re-calibrating the model
is recommended, especially when $\alpha$ is small.

%%%%%%%%%%%%%%%%%%%%%%%%%%%%%%%%%%%%%%%%%%%%%%%%%%%%%%%%%%%%
\subsection{Thinning Preserves the Bayes Decision Boundary}
\label{sec:bayes}

We can easily implement the thinning procedure, but
how will it affect the accuracy of the classifier?
The following result gives us a first promising piece of the answer by establishing
conditions under which thinning does not affect the Bayes decision boundary.

At a high level, our results rely on the fact that under our generative model,
the ``amount of information'' contained
in the input vector $X$ is itself uninformative about the class label $Y$.

\begin{assumption}[Equal information content across topics]\label{assu:indep_T}
Assume there exists a constant $\psi_0$ such that $\psi(\theta) = \psi_0$ with probability $1$,
over random $\theta$.
\end{assumption}
For example, in our Poisson topic model,
we imposed the restriction that $\psi(\theta) = \sum_{j=1}^d e^{\theta_j} = 1$,
which ensures that the document length $\|A_t\|_1$ has the same distribution
(which has expectation $\psi(\theta)$ in this case) for all possible $\theta$.

\begin{theorem}
\label{theo:bayes_boundary}
Under Assumption \ref{assu:indep_T},
the posterior class probabilities are invariant under thinning \eqref{eq:thinning}:
\begin{align}
  \label{eqn:bayes_boundary}
\PP{Y = y \cond \tX = x} = \PP{Y = y \cond X = x}
\end{align}
for all $y \in \{ 1, \dots, K \}$ and $x \in \mathcal{X}$.
\end{theorem}

\begin{proof}
  Given Assumption~\ref{assu:indep_T},
  the density of $A_t \mid \theta$ is given by:
  \begin{align}
    f_\theta^{(t)}(x) = e^{\theta \cdot x} e^{-t \psi_0} h^{(t)}(x),
  \end{align}
  which importantly splits into two factors, one depending on $(\theta,x)$, and the other depending on $(t,x)$.
  Now, let us compute the posterior distribution:
  \begin{align}
  \PP{Y = y \cond A_t = x}
  & \propto \PP{Y = y} \int \PP{\theta \mid Y} f_\theta^{(t)}(x) d\theta \\
  & \propto \PP{Y = y} \int \PP{\theta \mid Y} e^{\theta \cdot x} d\theta,
  \end{align}
  which does not depend on $t$, as $e^{-t \psi_0} h^{(t)}(x)$
  can be folded into the normalization constant.
  Recall that $X = A_T$ and $\tX = A_{\tT}$.
  Substitute $t = T$ and $t = \tT$ to conclude \refeqn{bayes_boundary}.
\end{proof}

To see the importance of Assumption~\ref{assu:indep_T},
consider the case where we have two labels ($Y \in \{ 1, 2 \}$),
each with a single topic ($Y$ yields topic $\theta_Y$).
Suppose that $\psi(\theta_2) = 2 \psi(\theta_1)$---that is,
documents in class 2 are on average twice as long as those in class 1.
Then, we would be able to make class 2 documents look like class 1 documents
by thinning them with $\alpha = 0.5$.

\begin{remark}\label{rem:obs_T2}
If we also condition on the information content $T$, then an analogue to Theorem~\ref{theo:bayes_boundary} holds even without Assumption~\ref{assu:indep_T}:
\begin{equation}
\PP{Y = y \cond \tX = x, \, \tT = t} = \PP{Y = y \cond X = x, \, T = t}.
\end{equation}
This is because, after conditioning on $T$, the $e^{-t\psi(\theta)}$ term factors out of the likelihood.
\end{remark}

The upshot of Theorem~\ref{theo:bayes_boundary} is that thinning
will not induce asymptotic bias whenever an estimator produces
$\PP{Y = y \cond X = x}$ in the limit of infinite data ($n \rightarrow \infty$), i.e., 
if the logistic regression (Algorithm~\ref{proc:levy_thinning}) is well-specified.
Specifically, training either on original examples or thinned examples will both converge
to the true class-conditional distribution.
The following result assumes that the feature space $\mathcal{X}$ is discrete; the proof
can easily be generalized to the case of continuous features.

\begin{corollary}
\label{coro:consistent}
Suppose that Assumption \ref{assu:indep_T} holds, and
that the above multi-class logistic regression model is well-specified, i.e.,
$\PP{Y = y \cond X = x} \propto e^{\beta^{(y)} \cdot \phi(x)}$
for some $\beta$ and all $y = 1, \, ..., \, K$.
Then, assuming that $\PP{A_t = x} > 0$ for all $x \in \mathcal{X}$ and $t > 0$,
Algorithm~\ref{proc:levy_thinning} is consistent,
i.e., the learned classification rule converges to the Bayes classifier as $n \rightarrow \infty$.
\end{corollary}

\begin{proof}
At a fixed $x$, the population loss $\EE{\ell\p{\beta;\, X, \, Y \cond X = x}}$
is minimized by any choice of $\beta$ satisfying:
\begin{equation}
\label{eq:well_spec}
\frac{\exp\sqb{\beta^{(y)} \cdot \phi(x)}}{\sum_{k = 1}^K \exp\sqb{\beta^{(k)} \cdot \phi(x)}} = \PP{Y = y \cond X = x}
\end{equation}
for all $y = 1, \, ..., \, K$. Since the model is well-specified and by assumption
$\mathbb{P}[\tX = x] > 0$ for all $x \in \mathcal{X}$, we conclude that weight vector $\hbeta$ learned using
Algorithm~\ref{proc:levy_thinning} must satisfy asymptotically \eqref{eq:well_spec} for all $x \in \mathcal{X}$
as $n \rightarrow \infty$.
\end{proof}

\subsection{The End of the Path}
\label{sec:end}

As seen above, if we have a correctly specified logistic regression model,
then L\'evy thinning regularizes it without introducing any bias.
However, if the logistic regression model is misspecified,
thinning will in general induce bias, and
the amount of thinning presents a bias-variance trade-off.
The reason for this bias is that although
thinning preserves the Bayes decision boundary,
it changes the marginal distribution of the covariates $X$,
which in turn affects 
logistic regression's linear approximation to
the decision boundary.
Figure~\ref{fig:bias} illustrates this phenomenon in the case where $A_t$ is
a Brownian motion, corresponding to Gaussian feature noising;
\citet{wager2014altitude} provides a similar
example for the Poisson topic model.

\begin{figure}[t]
\centering
\includegraphics[width=0.7\textwidth]{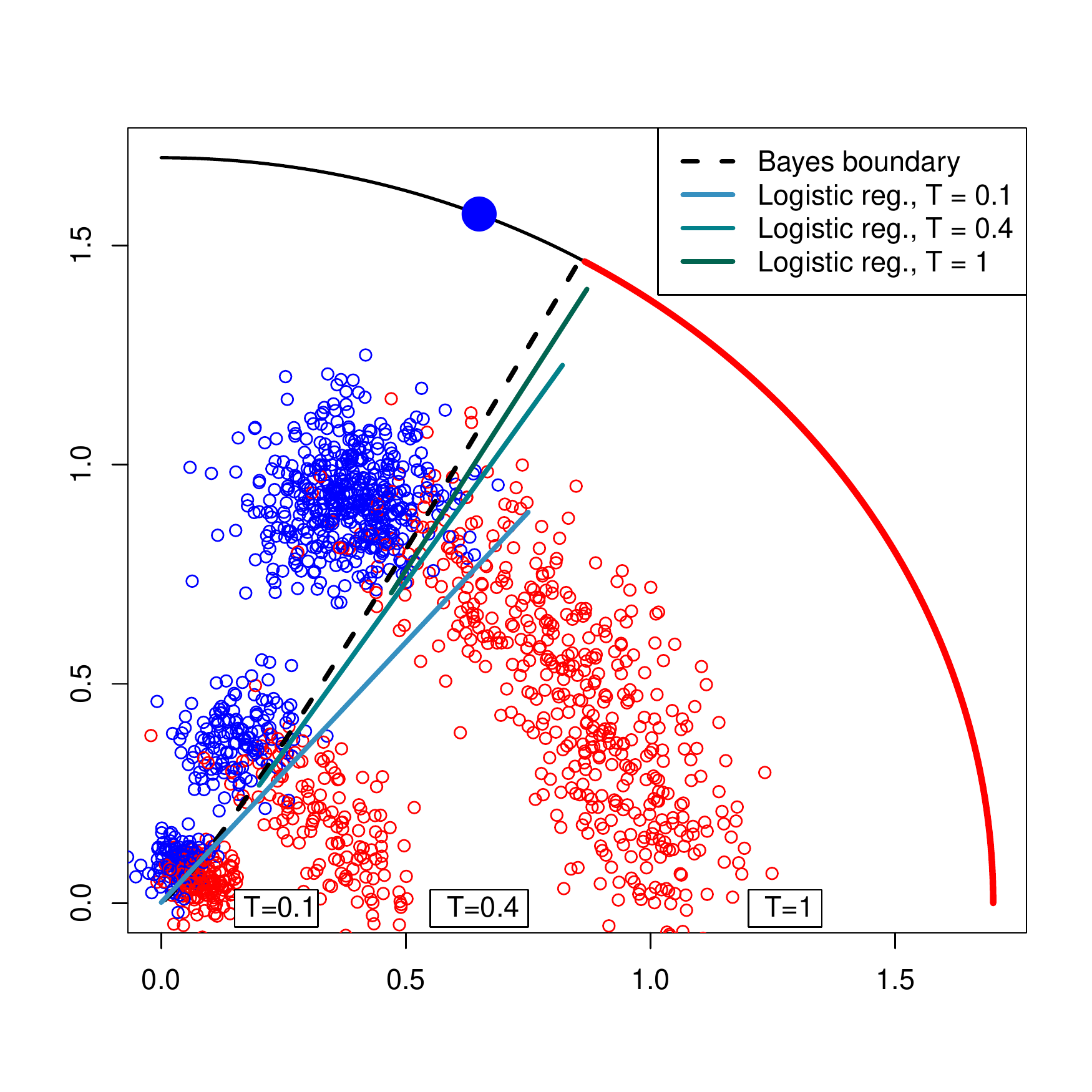}
\caption{The effect of L\'evy thinning with data generated from a Gaussian
model of the form $X \cond \theta, \, T \sim \mathcal{N}\p{T \, \theta,
\sigma^2 T \, I_{p \times p}}$, as described in Section \ref{sec:BM}. The outer
circle depicts the distribution of $\theta$ conditional on the color $Y$: blue
points all have $\theta \propto (\cos(0.75 \, \pi/2), \, \sin(0.75 \, \pi/2))$,
whereas the red points have $\theta \propto (\cos(\omega\, \pi/2), \,
\sin(\omega\, \pi/2))$ where $\omega$ is uniform between 0 and 2/3. Inside this
circle, we see 3 clusters of points generated with $T=0.1, \, 0.4, \, \eqand
1$, along with logistic regression decision boundaries obtained from each
cluster.
The dashed line shows the Bayes decision boundary
separating the blue and red points, which is the same for all $T$
(Theorem \ref{theo:bayes_boundary}).
Note that the logistic regression boundaries learned from data with different $T$ are not the
same. This issue arises because the Bayes decision boundary is curved, and the
best linear approximation to a curved Bayes boundary changes with $T$.}
\label{fig:bias}
\end{figure}

Fully characterizing the bias of L\'evy thinning is beyond the scope of
this paper. However, we can gain some helpful insights about this bias by
studying ``strong thinning''---i.e., L\'evy thinning in
the limit as the thinning parameter $\alpha \rightarrow 0$:
\begin{equation}
\label{eq:b0}
\hbeta_{0+} \eqdef \lim_{\alpha \rightarrow 0} \lim_{B \rightarrow \infty} \hbeta(\alpha, B),
\end{equation}
where $\hbeta(\alpha, B)$ is defined as in \eqref{eq:balpha} with the explicit dependence on $\alpha$ and $B$.
For each $\alpha$, we take $B \to \infty$ perturbed points for each of the
original $n$ data points.
As we show in this section, this limiting
classifier is well-defined under weak conditions;
moreover, in some cases of interest, it
can be interpreted as a simple generative classifier.
The result below concerns the existence of $\hbeta_{0+}$,
and establishes that it is the empirical minimizer of a convex loss function.

\begin{theorem}
\label{theo:strong_thin}
Assume the setting of Procedure \ref{proc:levy_thinning},
and let the feature map be $\phi(x) = x$.
Assume that the generative L\'evy process $(A_t)$ has finitely many
jumps in expectation over the interval $[0, T]$.
Then, the limit $\hbeta_{0+}$ is well-defined
and can be written as
\begin{equation}
\label{eq:balphastrong}
\hbeta_{0+} = \argmin_{\beta \in \R^{p \times K}} \cb{\sum_{i = 1}^n \limloss\p{\beta; \, X^{(i)}, \, Y^{(i)}}},
 \end{equation}
 for some convex function $\limloss(\cdot; \, x, \, y)$.
\end{theorem}
 
The proof of Theorem \ref{theo:strong_thin} is provided in the appendix. Here, we begin by
establishing notation that lets us write down an expression for the limiting loss $\limloss$.
First, note that Assumption \ref{assu:expfam} implicitly
requires that the process $(A_t)$ has finite moments. Thus, by the
L\'evy--It\=o decomposition, we can uniquely write this process as
\begin{equation}
\label{eq:decomp}
A_t = bt + W_t + N_t,
\end{equation}
where $b \in \RR^p$, $W_t$ is a Wiener process with covariance $\Sigma$, and $N_t$ is
a compound Poisson process which, by hypothesis, has a finite jump intensity.

Now, by an argument analogous to that in the proof of Theorem \ref{theo:marginal_thinning}, we
see that the joint distribution of $W_T$ and $N_T$ conditional on $A_T$ does not depend on $\theta$.
Thus, we can define the following quantities without ambiguity:
\begin{align}
\label{eq:muT}
&\mu_T(x) = bT + \EE{W_T \cond A_T = x},  \\
\label{eq:lambdaT}
&\lambda_T(x) = \EE{\text{number of jumps in $(A_t)$ for $t \in [0, \, T]$} \cond A_T = x}, \\
\label{eq:nuT}
&\nu_T(z; \, x) = \lim_{t \rightarrow 0} \PP{N_t = z \cond N_t \neq 0, \, A_T = x}.
\end{align}
More prosaically, $\nu_T(\cdot; \, x)$ can be described as the distribution of the first jump of $\tN_t$,
a thinned version of the jump process $N_t$.
In the degenerate case where $\PP{\N_T = 0 \cond A_T = x} = 0$, we set $\nu_T(\cdot; \, x)$ to be a point
mass at $z = 0$.

Given this notation,
we can write the effective loss function $\limloss$ for strong thinning as
\begin{align}
\label{eq:limloss}
\limloss\p{\beta; \, x, \, y}
&= -\mu_T(x) \cdot \beta^{(y)} + \frac{T}{2} \frac{1}{K} \sum_{k = 1}^K \beta^{(k)\top} \Sigma \beta^{(k)} \\
\notag
&\hspace{-2em} + \lambda_T(x) \int \ell\p{\beta; \, z, \, y} \ d\nu_T(z; \, x),
\end{align}
provided we require without loss of generality that $\sum_{k = 1}^K \beta^{(k)} = 0$.
In other words, the limiting loss can be described entirely in terms of the distribution of the first jump
of $\tN_t$, and continuous part $W_t$ of the L\'evy process. The reason for this phenomenon is that, in the
strong thinning limit, the pseudo-examples $\tX \sim A_{\alpha T}$ can all be characterized using
either 0 or 1 jumps.

Aggregating over all the training examples,
we can equivalently write this strong thinning loss as
\begin{align}
\notag
  \sum_{i = 1}^n \limloss\p{\beta; \, X^{(i)}, \, Y^{(i)}}  &= \frac{1}{2T} \sum_{i = 1}^n \gamma_{Y^{(i)}}^{-1} \Norm{\gamma_{Y^{(i)}} \, \mu_T\p{X^{(i)}} - T \Sigma \beta^{(Y^{(i)})}}_{\Sigma^{-1}}^2  \\
&\hspace{-2em}  + \sum_{i = 1}^n \lambda_T(X^{(i)}) \int \ell\p{\beta; \, z, \, Y^{(i)}} \ d\nu_T(z; \, X^{(i)}),
\end{align}
up to $\Norm{\mu_T}_2^2$ terms that do not depend on $\beta$.
Here, $\frac12 \Norm{v}_{\Sigma^{-1}}^2 = \frac12 v' \Sigma^{-1} v$ corresponds to the
Gaussian log-likelihood with covariance $\Sigma$ (up to constants),
and $\gamma_y = K  \abs{\cb{i : Y^{(i)} = y}} / n$ measures the over-representation
of class $y$ relative to other classes.

In the case where we have the same number of training examples from each class
(and so $\gamma_y = 1$ for all $y = 1, \, ..., \, K$),
the strong thinning loss can be understood in terms of a generative model. The first term, namely
$\frac{1}{2T} \sum_{i = 1}^n \Norm{\mu_T(X^{(i)}) - T\Sigma \beta^{(Y^{(i)})}}_{\Sigma^{-1}}^2, $
is the loss function for linear classification in a Gaussian mixture with observations $\mu_T(X^{(i)})$, while
the second term is the logistic loss obtained by classifying single jumps.
Thus, strong thinning is effectively seeking the best linear classifier for a generative model that is a
mixture of Gaussians and single jumps.

In the pure jump case ($\Sigma = 0$), we also note that strong thinning
is closely related to naive Bayes classification. In fact, if the jump measure of $N_t$
has a finite number of atoms that are all linearly independent, then we can verify that the
parameters $\hbeta_{0+}$ learned by strong thinning are equivalent to those learned
via naive Bayes, although the calibration constants $\ck$ may be different.

At a high level, by elucidating the generative model that strong thinning pushes us towards,
these results can help us better understand the behavior of L\'evy thinning for intermediate
value of $\alpha$, e.g., $\alpha = 1/2$.
They also suggest caution with respect to calibration: For both the diffusion and jump terms,
we saw above that L\'evy thinning gives helpful guidance for the angle of $\beta^{(k)}$, but
does not in general elegantly account for signal strength $\Norm{\beta^{(k)}}_2$ or relative
class weights. Thus, we recommend re-calibrating the class decision boundaries obtained
by L\'evy thinning, as in Algorithm~\ref{proc:levy_thinning}.

\section{Examples}\label{sec:examples}

So far, we have developed our theory of L\'evy thinning using the Poisson
topic model as a motivating example, which corresponds to dropping out words
from a document.  In this section, we
present two models based on other L\'evy processes---multivariate Brownian
motion (\refsec{BM}) and Gamma processes (\refsec{gamma})---
exploring the consequences of L\'{e}vy thinning.

\subsection{Multivariate Brownian Motion}
\label{sec:BM}

Consider a classification problem where the input vector is
the aggregation of multiple noisy, independent measurements
of some underlying object.
For example, in a biomedical application,
we might want to predict a patient's disease status based on a set of
biomarkers such as gene expression levels or brain activity.
A measurement is typically obtained through a
noisy experiment involving an microarray or fMRI,
so multiple experiments might be performed and aggregated.

More formally,
suppose that patient $i$ has disease status $\Yi$ and expression
level $\mu_i \in \R^d$ for $d$ genes,
with the distribution of $\mu_i$ different for each disease status.
Given $\mu_i$, suppose the $t$-th measurement for patient $i$
is distributed as
\begin{align}
Z_{i,t} \sim \mathcal N(\mu_i, \Sigma),
\end{align}
where $\Sigma \in \R^{d \times d}$ is assumed to be a known, fixed matrix.
Let the observed input be $\Xi= \sum_{t=1}^{T_i} Z_{i,t}$, the sum of the noisy
measurements. If we could take infinitely many measurements ($T_i\to\infty$),
we would have $\Xi/T_i \to \mu_i$ almost surely; that is, we would observe gene
expression noiselessly. For finitely many measurements, $\Xi$ is a noisy proxy
for the unobserved $\mu_i$.

We can model the process of accumulating measurements with a multivariate
Brownian motion $(A_t)$:
\begin{align}
A_t = t \mu + \Sigma^{1/2}B_t,
\end{align}
where $B_t$ is a $d$-dimensional white Brownian motion.\footnote{
By definition of Brownian motion,
we have marginally that $B_t \sim \mathcal N(0, tI)$.}
For integer values of $t$, $A_t$ represents the sum of the first $t$
measurements, but $A_t$ is also defined for fractional values of $t$. The
distribution of the features $X$ at a given time $T$ is thus
\begin{align}
X \mid \mu, T \sim \mathcal N(T\mu, T\Sigma),
\end{align}
leading to density
\begin{align}
  f_{\mu}^{(t)}(x)
  &= \frac{\exp\sqb{\frac{1}{2}(x-t\mu)^\top (t\Sigma)^{-1}(x - t\mu)}}{(2\pi)^{d/2}\det(\Sigma)}\\
  &= \exp\sqb{x^\top \Sigma^{-1}\mu - \frac{t}{2}\mu^\top \Sigma^{-1}\mu} h^{(t)}(x), \nonumber
\end{align}
where
\begin{align}
h^{(t)}(x) = 
\frac{\exp\sqb{-\frac{1}{2t} x^\top \Sigma^{-1}x}}
{(2\pi)^{d/2}\det(\Sigma)^{1/2}}.
\end{align}
We can recover the form of~\eqref{eq:assu1} by setting $\theta=\Sigma^{-1}\mu$,
a one-to-one mapping provided $\Sigma$ is positive-definite.

\paragraph{Thinning.}
The distribution of $\tX=A_{\alpha T}$ given $X=A_T$ is that of a Brownian bridge process
with the following marginals:
\begin{align}
\tX \mid X \sim \mathcal N\p{\alpha X, \alpha(1-\alpha)T\Sigma}.
\end{align}

In this case, ``thinning'' corresponds exactly to adding zero-mean, additive
Gaussian noise to the scaled features $\alpha X$. Note that in this model,
unlike in the Poisson topic model, sampling $\tX$ from $X$ does require
observing $T$---for example, knowing how many observations were taken. The
larger $T$ is, the more noise we need to inject to achieve the same
downsampling ratio.

In the Poisson topic model, the features $(X_{i,1},\ldots,X_{i,d})$ were
independent of each other given the topic $\theta_i$ and expected length $T_i$.
By contrast, in the Brownian motion model the features are correlated
(unless $\Sigma$ is the identity matrix). This serves to illustrate that
independence or dependence of the features is irrelevant to our general
framework; what is important is that the {\em increments} $Z_t = A_{t}-A_{t-1}$
are independent of each other, the key property of a L\'{e}vy process.

Assumption~\ref{assu:indep_T} requires that
$\mu^\top \Sigma^{-1}\mu$ is constant across topics; i.e., that the
true gene expression levels are equally sized in the Mahalanobis norm defined
by $\Sigma$.
Clearly, this assumption is overly stringent in real situations.
Fortunately, Assumption~\ref{assu:indep_T} is not required (see Remark \ref{rem:obs_T2})
as long as $T$ is observed---as it must be anyway if we want to be able to carry
out L\'evy thinning.

Thinning $X$ in this case is very similar to subsampling. Indeed, for integer
values of $\tT$, instead of formally carrying out L\'evy thinning as detailed above,
we could simply resample $\tT$ values of $Z_{i,t}$ without replacement,
and add them together to obtain $\tX$. If there are relatively few
repeats, however, the resampling scheme can lead to only $\binom{T}{\tT}$
pseudo-examples (e.g. 6 pseudo-examples if $T=4$ and $\tT=2$), whereas the
thinning approach leads to infinitely many possible pseudo-examples we can use
to augment the regression.
Moreover, if $T=4$ then subsampling leaves us with
only four choices of $\alpha$; there would be no way to thin using
$\alpha=0.1$, for instance.

\subsection{Gamma Process}
\label{sec:gamma}

As another example, suppose again that we are predicting a patient's disease
status based on repeated measurements of a biomarker such as gene expression or
brain activity. But now, instead of (or in addition to) the {\em average}
signal, we want our features to represent the variance or covariance of the
signals across the different measurements.

Assume first that the signals at different genes or brain locations
are independent; that is, the $t$-th measurement for patient $i$ and gene $j$
has distribution
\begin{align}
Z_{i,j,t} \sim \mathcal N(\mu_{i,j},\sigma_{i,j}^2).
\end{align}
Here, the variances $\sigma_i^2 = (\sigma_{i,1}^2,\ldots,\sigma_{i,d}^2)$
parameterize the ``topic.''
Suppressing the subscript $i$, after $T+1$ measurements we can compute
\begin{align}
X_{j,T} = \sum_{t=1}^{T+1} (Z_{i,j,t}- \bar Z_{i,j,T+1})^2, \quad \text{ where } \quad
\bar Z_{i,j,T+1} = \frac{1}{T+1} \sum_{t=1}^{T+1} Z_{i,j,t}.
\end{align}
Then $X_{j,T} \sim \sigma_j^2 \chi_T^2$, which is a Gamma distribution with
shape parameter $T/2$ and scale parameter $2\sigma_j^2$ (there is no dependence
on $\mu_i$). Once again, as we accumulate more and more observations (increasing $T$), we will
have $X_T/T \to (\sigma_1^2,\ldots,\sigma_d^2)$ almost surely.

We can embed $X_{j,T}$ in a multivariate Gamma process with $d$ independent coordinates and scale parameters $\sigma_j^2$:
\begin{align}
(A_t)_j \sim \text{Gamma}(t/2,2\sigma_j^2).
\end{align}
The density of $A_t$ given $\sigma^2$ is
\begin{align}
f_{\sigma^2}^{(t)}(x) 
&= \prod_{j=1}^d \frac{x_j^{t/2-1}e^{-x_j/2\sigma_j^2}}{\Gamma(t/2)2^{t/2}\sigma_j^{2 (t/2)}} \\
&= \exp\sqb{ -\sum_{j=1}^d x_j/2\sigma_j^2 - (t/2)\sum_{j=1}^d \log \sigma_j^2} h^{(t)}(x), \nonumber
\end{align}
where
\begin{align}
h^{(t)}(x) = \frac{\prod_j x_j^{t/2-1}}{\Gamma(t/2)^d 2^{dt/2}}.
\end{align}
We can recover the form of~\eqref{eq:assu1} by setting
$\theta_j=-1/2\sigma_j^2$, a one-to-one mapping.

\paragraph{Thinning.}
Because $\tX_j\sim \text{Gamma}(\alpha T/2,2\sigma_j^2)$ is independent of the increment
${X_j-\tX_j\sim \text{Gamma}((1-\alpha)T/2, 2\sigma_j^2)}$, we have
\begin{align}
\frac{\tX_j}{X_j} \mid X_j \sim \text{Beta}\p{\alpha T/2, (1-\alpha)T/2}.
\end{align}
In other words, we create a noisy $\tX$ by generating for each coordinate an independent {\em multiplicative} noise factor
\begin{align}
m_j \sim \text{Beta}\p{\alpha T, (1-\alpha)T}
\end{align}
and setting  $\tX_j = m_jX_j$. Once again, we can downsample without knowing
$\sigma_j^2$, but we do need to observe $T$.
Assumption~\ref{assu:indep_T} would
require that $\prod_j \sigma_j^2$ is identical for all topics.
This is an unrealistic assumption,
but once again it is unnecessary as long as we observe $T$.

\paragraph{General covariance.}

More generally, the signals at different brain locations, or expressions for
different genes, will typically be correlated with each other, and these
correlations could be important predictors.
To model this, let the measurements be distributed as:
\begin{align}
Z_{i,t} \sim \mathcal N(\mu_i,\Sigma_i),
\end{align}
where $\Sigma$ represents the unknown ``topic''---some covariance matrix that is
characteristic of a certain subcategory of a disease status.

After observing $T+1$ observations we can construct the matrix-valued features:
\begin{align}
X_{T} = \sum_{t=1}^{T+1} (Z_{i,t}- \bar Z_{i,T+1}) (Z_{i,t} - \bar Z_{i,T+1})^\top.
\end{align}
Now $X_T$ has a Wishart distribution: $X_T \sim \text{Wish}_d(\Sigma,T)$.
When $T\geq d$, the density of $A_t$ given $\Sigma$ is
\begin{align}
f_{\Sigma}^{(t)}(x) 
  = \exp\left\{
    -\frac{1}{2} \tr(\Sigma^{-1} x)
    -\frac{t}{2} \log \det(\Sigma)
    \right\}h^{(t)}(x),
\end{align}
where
\begin{align}
  h^{(t)}(x) &= \left(2^{\frac{td}{2}} \det(x)^{\frac{t-d-2}{2}}\Gamma_d\left(\frac{t}{2}\right)\right)^{-1}, \\
  \Gamma_d\left(\frac{t}{2}\right) &= \pi^{\frac{d(d-1)}{4}} \prod_{j=1}^d \Gamma\left(\frac{t}{2} + \frac{1-j}{2}\right),
\end{align}
supported on positive-definite symmetric matrices. If $X=A_T$ and $\alpha T \geq d$ as well, we can sample a ``thinned'' observation $\tX$ from density proportional to
\begin{align}
h^{(\alpha T)}(\tilde x) h^{(T-\alpha T)}(X-\tilde x) \propto \det(\tilde x)^{\frac{2+d-\alpha T}{2}} \det(X-\tilde x)^{\frac{2+d-(1-\alpha) T}{2}},
\end{align}
or after the affine change of variables $\tX = X^{1/2}MX^{1/2}$, we sample $M$
from density proportional to $\det(m)^{\frac{2+d-\alpha
T}{2}}\det(I_d-m)^{\frac{2+d-(1-\alpha)t}{2}}$, a matrix beta distribution.
Here, $M$ may be interpreted as matrix-valued multiplicative noise.

\section{Simulation Experiments}
\label{sec:simu}

\begin{figure}[t]
\centering
\includegraphics[width=0.7\textwidth]{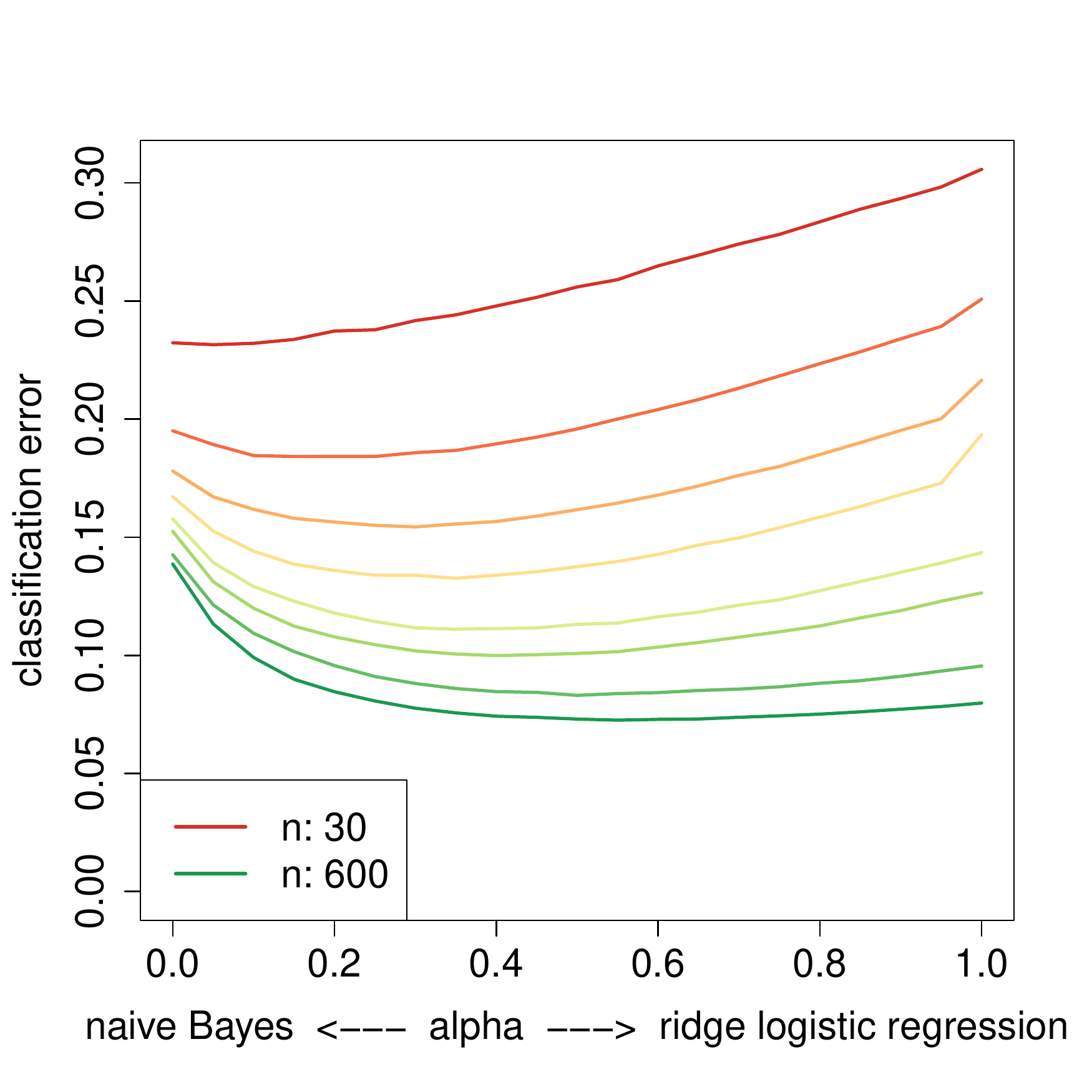}
\caption{Performance of L\'evy thinning with cross-validated ridge-regularized logistic regression,
on a random Gaussian design described in \eqref{eq:gauss_simu}. The curves depict the relationship
between thinning $\alpha$ and classification error as
the number of training examples grows:
$n = 30, \, 50, \,  75, \,  100, \,  150, \, 200, \,  400, \eqand 600$.
We see that naive Bayes improves over ridge logistic regression in very small samples,
while in moderately small samples L\'evy thinning does better than either end of the path.
\label{fig:gauss_simu}
}
\end{figure}

\begin{figure}[t]
\centering
\includegraphics[width=0.7\textwidth]{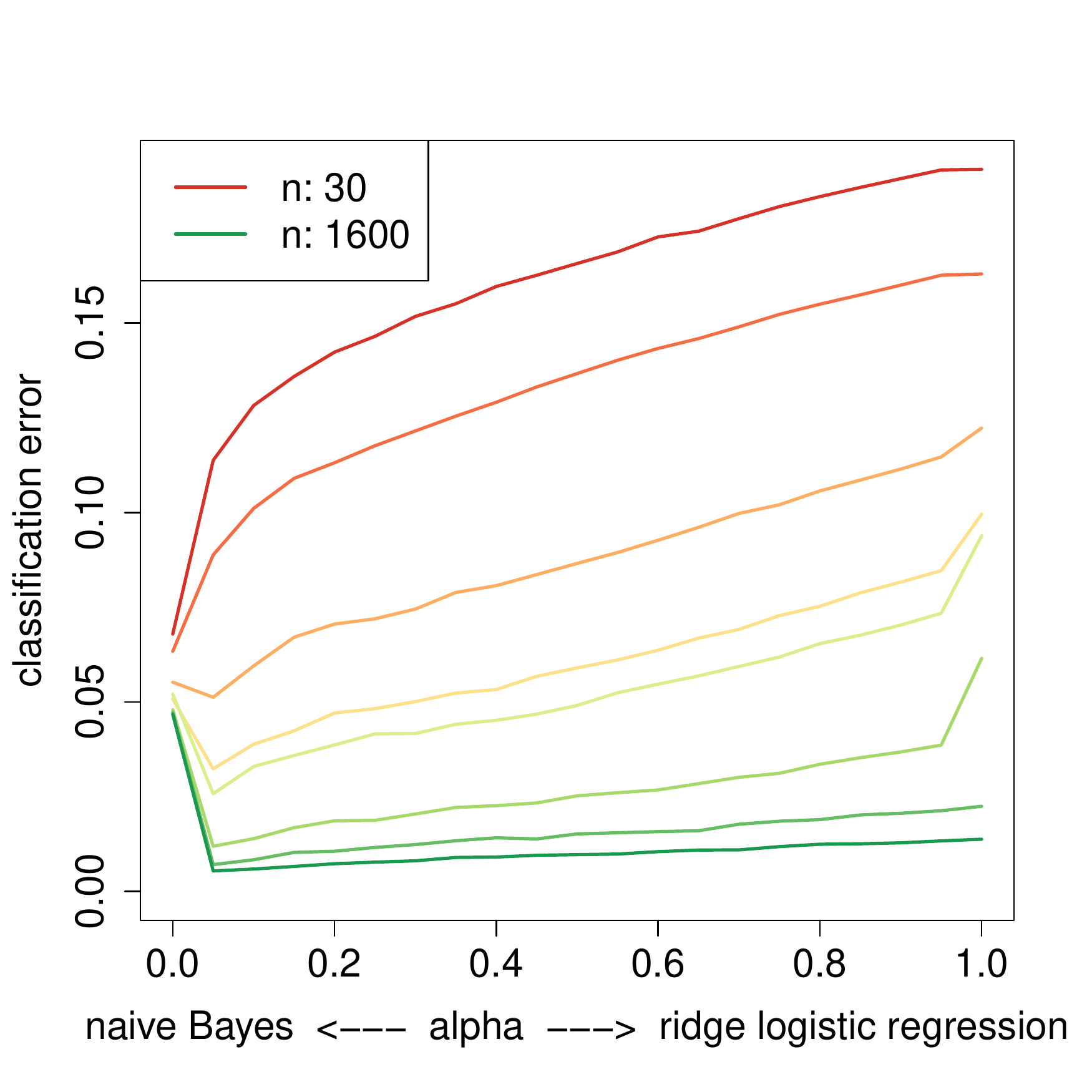}
\caption{Performance of L\'evy thinning with cross-validated ridge-regularized logistic regression,
on a random Poisson design described in \eqref{eq:poisson_simu}. The curves depict the relationship
between thinning $\alpha$ and classification accuracy for
$n = 30, \, 50, \,  100, \,  150, \, 200, \,  400, \, 800, \eqand 1600$.
Here, aggressive L\'evy thinning with small but non-zero $\alpha$ does substantially better than
naive Bayes ($\alpha = 0$) as soon as $n$ is moderately large.
\label{fig:poisson_simu}
}
\end{figure}

In this section,
we perform several simulations to illustrate
the utility of L\'evy thinning.
In particular,
we will highlight the modularity between L\'evy thinning
(which provides pseudo-examples)
and the discriminative learner (which ingests these pseudo-examples).
We treat the discriminative learner as a black box,
complete with its own internal cross-validation scheme that
optimizes accuracy on pseudo-examples.
Nonetheless, we show that accuracy on the original examples
improves when we train on thinned examples.

More specifically, given a set of training examples $\{ (X,Y) \}$,
we first use L\'evy thinning to
generate a set of pseudo-examples $\{ (\tX, \, Y) \}$.
Then we feed these examples to the
\texttt{R} function \texttt{cv.glmnet} to learn a linear classifier
on these pseudo-examples \citep{friedman2010regularization}.
We emphasize that  \texttt{cv.glmnet} seeks to choose its
regularization parameter $\lambda$ to maximize its accuracy on the pseudo-examples $(\tX, \, Y)$ rather than
on the original data $(X, \, Y)$.
Thus, we are using cross-validation as a black box
instead of trying to adapt the procedure to the context of L\'evy thinning.
In principle, we might be concerned that cross-validating
on the pseudo-examples would yield a highly suboptimal choice of $\lambda$, but our experiments
will show that the procedure in fact works quite well.

The two extremes of the path correspond to naive
Bayes generative modeling at one end ($\alpha = 0$), and plain ridge-regularized logistic regression
at the other ($\alpha = 1$).
All methods were calibrated on the training data as follows: Given
original weight vectors $\hbeta$, we first compute un-calibrated predictions $\hat\mu = X\hbeta$
for the log-odds of $\PP{Y = 1 \cond X}$, and then run a second univariate logistic regression
$Y \sim \hat\mu$ to adjust both the intercept and the magnitude of the original coefficients.
Moreover, when using cross-validation on pseudo-examples $(\tX, \, Y)$, we ensure that all
pseudo-examples induced by a given example $(X, \, Y)$ are in the same cross-validation fold.
Code for reproducing our results is available at \texttt{https://github.com/swager/levythin}.

\paragraph{Gaussian example.}
We generate data from the following hierarchical model:
\begin{equation}
\label{eq:gauss_simu}
Y \sim \text{Binomial}\p{0.5}, \ \ \mu \cond Y \sim \law_Y, \ \ X \cond \mu \sim \mathcal{N}\p{\mu, \, I_{d \times d}},
\end{equation}
where $\mu, \, X \in \RR^d$ and $d = 100$. The distribution $\law_Y$ associated with each label
$Y$ consists of 10 atoms $\mu^{(Y)}_1, \, ..., \, \mu^{(Y)}_{10}$. These atoms themselves are all
randomly generated such that their first 20 coordinates are independent draws of $1.1 \,  T_4$ where
$T_4$ follows Student's $t$-distribution with 4 degrees of freedom; meanwhile, the last 80 coordinates
of $\mu$ are all 0.
The results in Figure \ref{fig:gauss_simu} are marginalized over the randomness in $\law_Y$;
i.e., different simulation realizations have different conditional laws for $\mu$ given $Y$.
Figure~\ref{fig:gauss_simu} shows the results.

\paragraph{Poisson example.}
We generate data from the following hierarchical model:
\begin{equation}
\label{eq:poisson_simu}
Y \sim \text{Binomial}\p{0.5}, \ \ \theta \cond Y \sim \law_Y, \ \ X_j \cond \theta \sim \Pois\p{1000 \frac{e^{\theta_j}}{\sum_{j = 1}^d e^{\theta_j}}},
\end{equation}
where $\theta \in \RR^d$, $X \in \NN^d$, and $d = 500$. This time, however, $\law_Y$ is
deterministic: If $Y = 0$, then $\theta$ is just 7 ones followed by 493 zeros, whereas
$$ \theta \cond Y = 1 \sim \p{\underbrace{0, \, ..., \, 0}_7 \cond \underbrace{\tau, \, ..., \, \tau}_7 \cond \underbrace{0, \, ..., \, 0}_{486}}, \ \ \text{with} \ \ \tau \sim \Exp(3). $$
This generative model was also used in simulations by \citet{wager2014altitude}; the difference is that they applied
thinning to plain logistic regression, whereas here we verify that L\'evy thinning is also helpful when
paired with cross-validated ridge logistic regression.
Figure~\ref{fig:poisson_simu} shows the results.

These experiments suggest that it is reasonable to pair L\'evy
thinning with a well-tuned black box learner on the pseudo-examples $(\tX, \, Y)$,
without worrying about potential interactions between
L\'evy thinning and the tuning of the discriminative model.

\section{Discussion}

% Summary
In this chapter, we have explored a general framework for performing data
augmentation: apply L\'evy thinning and train a discriminative
classifier on the resulting pseudo-examples.  The exact thinning scheme reflects
our generative modeling assumptions.  We emphasize that the generative assumptions
are non-parametric and of a structural nature; in particular, we never fit an actual generative model,
but rather encode the generative hints implicitly in the pseudo-examples.

% Theoretical questions on current problem; Characterizing the bias
A key result is that under the generative assumptions, thinning preserves the
Bayes decision boundary, which suggests that a well-specified classifier incurs
no asymptotic bias.  Similarly, we would expect that a misspecified but powerful
classifier should incur little bias.
We showed that in limit of maximum thinning,
the resulting procedure corresponds to fitting a generative model.
The exact bias-variance trade-off for moderate levels of thinning
is an interesting subject for further study.

% Levy processes not the only structure
% Other forms of data augmentation
While L\'evy processes provide a general framework for thinning examples,
we recognize that there are many other forms of coarsening that could lead
to the same intuitions.  For instance, suppose $X \mid \theta$ is a Markov process over words
in a document.  We might expect that short \emph{contiguous} subsequences
of $X$ could serve as good pseudo-examples.  More broadly,
there are many forms of data augmentation that do not have the intuition of coarsening
an input.  For example, rotating or shearing an image to generate pseudo-images appeals
to other forms of transformational invariance.  It would be enlightening
to establish a generative framework in which data augmentation with these other forms of invariance
arise naturally.

\section{Appendix: Proof of Theorem \ref{theo:strong_thin}}

To establish the desired result, we show that for a single training example
$(X, \, Y)$, the following limit is well-defined for any $\beta \in \RR^{p \times K}$:
\begin{align}
\label{eq:limit}
\limloss\p{\beta; \, X, \, Y}
&= \lim_{\alpha \rightarrow 0} \frac{1}{\alpha} \p{\tEE{\ell\p{\beta; \, \tX, \, Y}} - \log\p{K}} \\
\notag
&\hspace{-2em}= -\beta^{(Y)} \cdot X + \lim_{\alpha \rightarrow 0} \frac{1}{\alpha} \tEE{\log\p{\frac{1}{K} \sum_{k = 1}^K e^{\beta^{(k)} \cdot \tX}}},
\end{align}
where on the second line we wrote down the logistic loss explicitly and
exploited linearity of the term involving $Y$ as in \citet{wager2013dropout}.
Here $\widetilde{\mathbb{E}}$ denotes expectation with respect to the thinning process
and reflects the $B \rightarrow \infty$ limit. Because $\ell$ is convex, $\limloss$ must also be
convex; and by equicontinuity $\hbeta(\alpha)$ must also converge to its minimizer.

Our argument relies on the decomposition $A_t = bt + W_t + N_t$ from \eqref{eq:decomp}.
Without loss of generality, we can generate the pseudo-features $\tX$ as
$\tX = bt + \tW_{\alpha T} + \tN_{\alpha T}$, where $\tW_{\alpha T}$ and $\tN_{\alpha T}$
have the same marginal distribution as $W_{\alpha T}$ and $N_{\alpha T}$.
Given this notation,
\begin{align*}
\frac{1}{\alpha} &\tEE{\log\p{\frac{1}{K} \sum_{k = 1}^K e^{\beta^{(k)} \cdot \p{\alpha b T + \tW_{\alpha T} + \tN_{\alpha T}}}}} \\
&= \frac{1}{\alpha} \tEE{\log\p{\frac{1}{K} \sum_{k = 1}^K e^{\beta^{(k)} \cdot \p{\alpha b T + \tW_{\alpha T}}}} \cond \tN_{\alpha T} = 0} \PP{\tN_{\alpha T} = 0}\\
&\hspace{2em} + \frac{1}{\alpha} \tEE{\log\p{\frac{1}{K} \sum_{k = 1}^K e^{\beta^{(k)} \cdot \p{\alpha b T + \tW_{\alpha T} + \tN_{\alpha T}}}} \cond \tN_{\alpha T} \neq 0} \PP{\tN_{\alpha T} \neq 0}.
\end{align*}
We now characterize these terms individually.
First, because $N_t$ has a finite jump intensity, we can verify that, almost surely,
$$ \lim_{\alpha \rightarrow 0} \frac{1}{\alpha} \PP{\tN_{\alpha T} \neq 0} = \lambda_T(X), $$
where $\lambda_T(X)$ is as defined in \eqref{eq:lambdaT}.
Next, because $\tW_{\alpha T}$ concentrates at 0 as $\alpha \rightarrow 0$, we can
check that
\begin{align*}
\lim_{\alpha \rightarrow 0} &\ \tEE{\log\p{\frac{1}{K} \sum_{k = 1}^K e^{\beta^{(k)} \cdot \p{\alpha b T + \tW_{\alpha T} + \tN_{\alpha T}}}} \cond \tN_{\alpha T} \neq 0} \\
&= \lim_{\alpha \rightarrow 0} \ \tEE{\log\p{\frac{1}{K} \sum_{k = 1}^K e^{\beta^{(k)} \cdot \tN_{\alpha T}}} \cond \tN_{\alpha T} \neq 0} \\
&=\int \log\p{\frac{1}{K} \sum_{k = 1}^K e^{\beta^{(k)} \cdot z}} \ d\nu_T(z; \, X)
\end{align*}
where $\nu_T(\cdot; \, X)$ \eqref{eq:nuT} is the first jump measure conditional on $X$.

Meanwhile, in order to control the remaining term, we note that we can write
$$ \tW_{\alpha T} = \alpha \tW_T + \tB_{\alpha T}, $$
where $\tB_t$ is a Brownian bridge from $0$ to $T$ that is independent from $\tW_T$.
Thus, noting that $\lim_{\alpha \rightarrow 0} \PP{\tN_{\alpha T} = 0} = 1$, we find that
\begin{align*}
\lim_{\alpha \rightarrow 0} \frac{1}{\alpha} &\tEE{\log\p{\frac{1}{K} \sum_{k = 1}^K e^{\beta^{(k)} \cdot \p{\alpha b T + \tW_{\alpha T}}}} \cond \tN_{\alpha T} = 0} \PP{\tN_{\alpha T} = 0} \\
&= \lim_{\alpha \rightarrow 0} \ \frac{1}{\alpha}\,  \tEE{\log\p{\frac{1}{K} \sum_{k = 1}^K e^{\beta^{(k)} \cdot \p{\alpha \p{bT + \tW_{T}}} + \tB_{\alpha T}}}} \\
&= \bar{\beta} \cdot \mu_T(X) + \frac{T}{2} \p{\frac{1}{K} \sum_{k = 1}^K \beta^{(k)\top} \Sigma \, \beta^{(k)} - \bar{\beta}^\top \Sigma \bar{\beta}},
\end{align*}
where $\mu_T(X)$ is as defined in \eqref{eq:muT} and $\bar{\beta} = K^{-1} \sum_{k = 1}^K \beta^{(k)}$.
The last equality follows from Taylor expanding the $\log(\sum \exp)$ term and noting that
3rd- and higher-order terms vanish in the limit.

Bringing back the linear term form \eqref{eq:limit},
and assuming without loss of generality that $\bar{\beta} = 0$,
we finally conclude that
\begin{align*}
\limloss\p{\beta; \, X, \, Y}
&= -\beta^{(Y)} \cdot X
+ \frac{T}{2} \, \frac{1}{K} \sum_{k = 1}^K \beta^{(k)\top} \Sigma \, \beta^{(k)} \\
&\hspace{-2em} + \lambda_T(X) \int \log\p{\frac{1}{K}\sum_{k = 1}^K e^{ \beta^{(k)} \cdot z}} \ d\nu_T(z; \, X) \\
&= -\beta^{(Y)} \cdot \mu_T(X)
+ \frac{T}{2} \, \frac{1}{K} \sum_{k = 1}^K \beta^{(k)\top} \Sigma \, \beta^{(k)} \\
&\hspace{-2em} + \lambda_T(X) \int - \beta^{(Y)} \cdot z  + \log\p{\sum_{k = 1}^K e^{ \beta^{(k)} \cdot z}} - \log(K) \ d\nu_T(z; \, X),
\end{align*}
where for the second equality we used the fact that $X = \mu_T(X) + \lambda_T(X)  \int z \ d\nu_T(z; \, X)$.
Finally, this expression only differs from \eqref{eq:limloss} by terms that do not include $\beta$; thus, they yield
the same minimizer.

{\small
\bibliographystyle{abbrvnat}
\bibliography{../refdb/all}
}

%%% Local Variables: 
%%% mode: latex
%%% TeX-master: "../asp_book"
%%% End: 

\endgroup

%%%%%%%%%%%%%%%%%%%%%%%%%%%%%%%%%%%%%%%%%%%%%%%%%%%%%%%%%%%%%%%%%%%%%
% TAIL                                                      %
%%%%%%%%%%%%%%%%%%%%%%%%%%%%%%%%%%%%%%%%%%%%%%%%%%%%%%%%%%%%%%%%%%%%%
%\printindex

\end{document}